\documentclass{article}

\PassOptionsToPackage{numbers,sort&compress}{natbib}
\usepackage[preprint]{neurips_2026}

\usepackage[utf8]{inputenc} 
\usepackage[T1]{fontenc}    
\usepackage{xcolor}
\definecolor{mydarkblue}{rgb}{0,0.08,0.45} 
\definecolor{darkgreen}{RGB}{3, 158, 23}
\usepackage[colorlinks,allcolors=mydarkblue,breaklinks]{hyperref}
\usepackage{url}            
\usepackage{booktabs}       
\usepackage{amsfonts}       
\usepackage{nicefrac}       
\usepackage{microtype}      
\usepackage{xcolor}         

\usepackage{caption}
\usepackage{enumitem}
\usepackage{multirow} 
\usepackage{booktabs}
\usepackage{arydshln}   
\usepackage{siunitx}
\usepackage{graphicx}
\usepackage{tikz}
\newlength{\figurewidth}
\usetikzlibrary{positioning}
\usepackage{amsmath}
\usepackage{amsthm}
\usepackage{graphicx}
\usepackage{amssymb}
\usepackage{colortbl}
\usepackage[capitalize,nameinlink]{cleveref}
\theoremstyle{definition}

\crefname{theorem}{Theorem}{Theorems}
\Crefname{theorem}{Theorem}{Theorems}
\usepackage{tabularx}
\usepackage{enumitem}

\renewcommand{\paragraph}[1]{\noindent\textbf{#1}~~}

\title{Point Tracking Improves World Action Models}

\author{
\textbf{Jiarui Guan}\textsuperscript{\normalfont 1} \quad
\textbf{Wenshuai Zhao}\textsuperscript{\normalfont 1,2,$\dagger$} \quad
\textbf{Yue Pei}\textsuperscript{\normalfont 6} \\
\textbf{Ziliang Chen}\textsuperscript{\normalfont 4,5} \quad
\textbf{Arno Solin}\textsuperscript{\normalfont 1,2} \quad
\textbf{Juho Kannala}\textsuperscript{\normalfont 1,3} \\
\textsuperscript{1}Aalto University \quad
\textsuperscript{2}ELLIS Institute Finland \quad
\textsuperscript{3}University of Oulu \\
\textsuperscript{4}Sun Yat-sen University \quad
\textsuperscript{5}Peng Cheng Laboratory \quad
\textsuperscript{6}Beihang University \\
{\tt\small \{jiarui.guan, wenshuai.zhao\}@aalto.fi} \\
{\tt\small \{juho.kannala, arno.solin\}@aalto.fi \quad c.ziliang@yahoo.com \quad peiyue@buaa.edu.cn} \\
\textsuperscript{$\dagger$}Corresponding author.
}
\begin{document}

\maketitle

\begin{abstract}

Robot policy learning benefits from world-action models that capture environment dynamics, but pixel-level prediction entangles dynamics with nuisance factors such as lighting and texture, making learned representations vulnerable to task-irrelevant visual variation. We propose JOPAT, a JOint Pixel-And-Track World-Action Model that predicts latent visual observations, 2D point tracks with visibility, and actions in a single denoising diffusion transformer. The key insight is that tracks provide an explicit representation of motion that captures long-horizon dynamics and remains robust under occlusion or partial out-of-frame motion, offering greater utility than modeling pixel appearance alone. On LIBERO and real-world LeRobot tasks, JOPAT improves over pixel-based baselines, with the largest gains on long-horizon tasks involving occlusion, object interaction, and off-screen motion.

\end{abstract}

\section{Introduction}
\label{sec:Intro}

Robotic manipulation in the open world demands policies that generalize across objects, scenes, and viewpoints while remaining reliable under partial observability, occlusion, and long-horizon temporal dependencies. 
Vision-Language-Action (VLA) models represent a major step toward this goal, which leverages pretrained Vision-Language Models (VLMs) to map language instructions and visual observations directly into robot actions ~\cite{black2026pi0visionlanguageactionflowmodel,brohan2022rt,cheang2025gr,kim2024openvla,shao2025large,shukor2025smolvla,team2024octo,zhang2025pure,zitkovich2023rt}. 
However, despite their strong semantic capabilities, VLAs are predominantly trained on image--language data and therefore lack an understanding of environment dynamics, which is crucial for robotic policy learning~\citep{wang2025vlatest,zhang2025vla,swann2026sparse,grover2025enhancing,li2026matters}.


Recently, World Action Models (WAMs) have emerged as a promising paradigm for translating video priors into robot control \cite{zhu2025unified,guo2026unified,chen2025unified,liu2025hybridvla,songdiva,won2025dual,wang2025unified,routray2025vipra,zhu2025wmpo,hegde2024warpd,zhang2025dreamvla,ye2026world,fan2026aim,zhang2025step,li2025comprehensive}.
Large-scale visual pre-training provides an effective way to acquire motion priors and physical grounding from diverse visual dynamics \cite{wu2023unleashing,agarwal2025cosmos,cheang2024gr,nair2022r3m,radosavovic2023robot,parisi2022unsurprising}. However, how to effectively translate video dynamics into generalizable robotic policies still remains an open question. Several studies suggest that pixel-level pretraining on action-free videos does not necessarily improve policy success rates~\cite{li2026matters,shiteleportation,rhoda2026dva,he2024learning}.

\begin{figure}[t]
    \centering
    \includegraphics[
        width=\textwidth,
        trim=5cm 0cm 0cm 0cm,
        clip
    ]{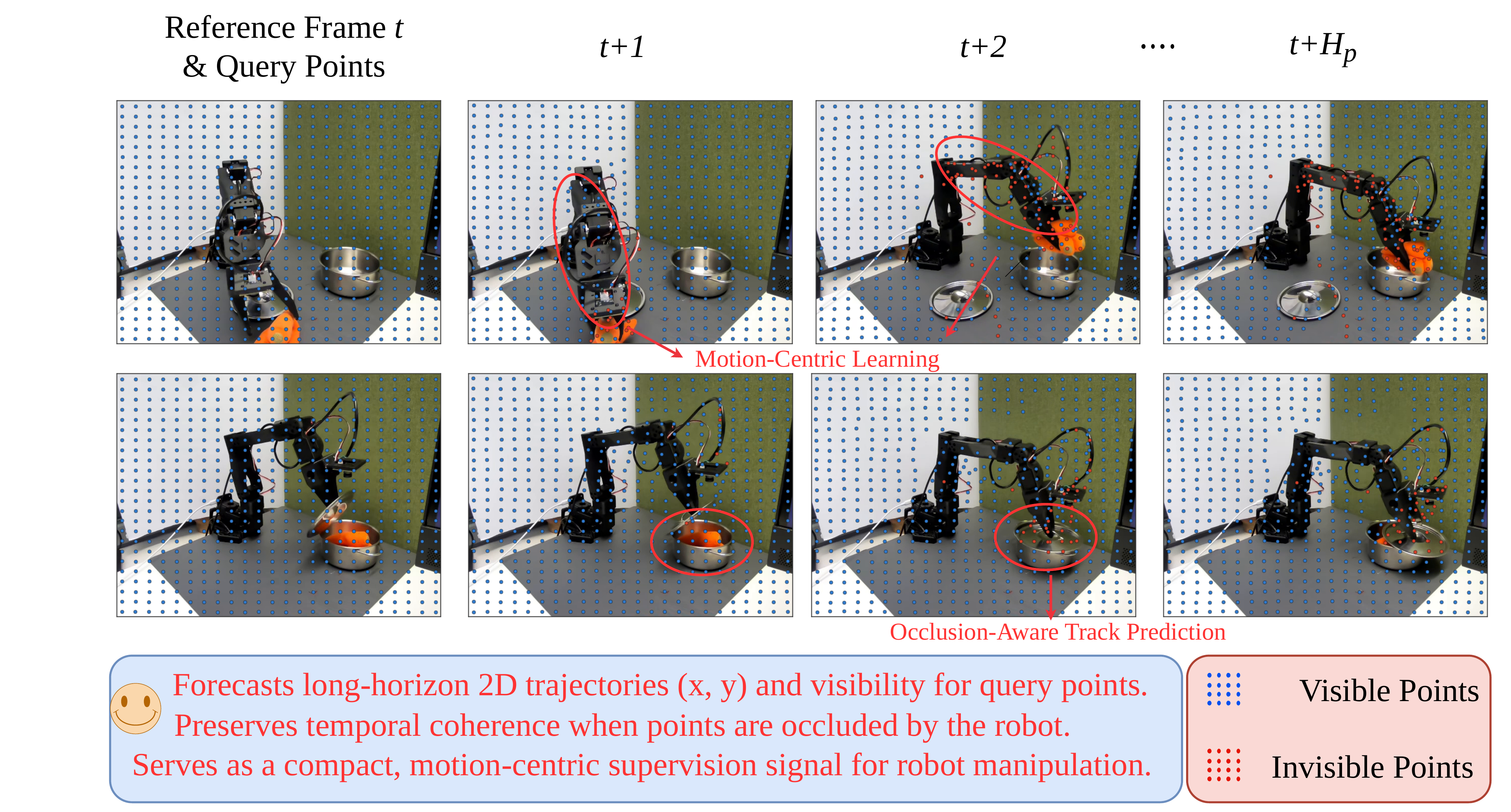}
    \vspace{-0.5em}
    \caption{
    \textbf{JOPAT predicts structured point tracks beyond visible pixels.}
    Starting from query points on the reference frame, JOPAT forecasts long-horizon 2D trajectories and visibility logits.
    This track-space prediction exposes action-relevant scene motion while explicitly representing points that become unobservable or leave the field of view.
    }
    \vspace{-0.8em}
    \label{fig:overview}
\end{figure}

We shift the central question from {\color{orange}\emph{how to model the world}} to {\color{darkgreen}\emph{what future state should serve as the interface between video prediction and control}}. Our key insight is that action-free video becomes useful for control when its supervision is expressed through a future-state abstraction that preserves visual grounding while exposing controllable correspondences. To this end, we propose the \emph{JOint Pixel-And-Track World-Action Model} (JOPAT), which jointly denoises visual latents, point tracks with visibility, and executable robot actions. In this way, we use 2D point tracks with visibility as the motion representation: tracks encode persistent scene correspondences over time, while visibility indicates when these correspondences become occluded or leave the camera view. Together with visual latents, this {\color{darkgreen}pixel–track future state} preserves object identity, affordance, and appearance cues, while explicitly representing object displacement, contact-induced motion, occlusion, and out-of-frame transitions, as illustrated in \cref{fig:overview}.

We evaluate JOPAT on both simulated and real-world manipulation benchmarks under visual out-of-distribution variations, self-induced occlusions, and long-horizon dependencies. On 40 LIBERO tasks~\cite{liu2023libero}, JOPAT achieves an average success rate of 97.8\%, establishing a new state of the art. It also demonstrates improved robustness on modified LIBERO settings with occlusions and disturbances. Real-world experiments and extensive ablation studies further validate the effectiveness of the proposed joint pixel-and-track modeling. Overall, integrating pixel-based visual representations with explicit point-track dynamics yields more reliable control than either signal alone, while enabling effective policy learning from limited robot demonstrations. 





Our main contributions are: (i) we propose JOPAT, a joint pixel-and-track world-action model that integrates visual latents, 2D point tracks with visibility, and robot actions within a unified generative framework, enabling correspondence-level motion to serve as an explicit interface between perception and control; (ii) we achieve state-of-the-art performance on 40 LIBERO manipulation tasks with an average success rate of 97.8\%, while demonstrating strong robustness under visual distribution shifts, occlusions, disturbances, and long-horizon dependencies in both simulated and real-world settings; and (iii) through extensive ablation studies and real-world experiments, we show that explicit point-track modeling is crucial for robust performance, as it improves long-horizon consistency, mitigates the effects of occlusion, and provides stable motion grounding that complements pixel-based representations for control.


\section{Related work}
\label{sec:related_work}

\paragraph{Video-based world models and world action models}
Video-based world models learn predictive dynamics from future observations and have long been used as a route to physical interaction and control~\cite{finn2016unsupervised,babaeizadeh2017stochastic,bruce2024genie}.
Recent robotic world-action models further couple video prediction with action generation, using shared generative models, diffusion objectives, or policy optimization to connect visual futures to executable controls~\cite{zhu2025unified,guo2026unified,routray2025vipra,zhu2025wmpo,hegde2024warpd,zhang2025dreamvla,ye2026world,fan2026aim}.
These methods show that predictive visual objectives can improve policy learning, but most instantiate the future state in pixel, latent-image, or visual-token spaces.
As a result, the motion variables that determine actions---which object moved, where it moved, and whether the evidence became hidden---must be recovered implicitly from appearance-dominated tokens.
JOPAT differs by augmenting the world-action future state with explicit point tracks and visibility, so action tokens interact with sampled correspondence-level motion variables rather than with visual latents alone.

\paragraph{Intermediate motion representations for control}
A growing line of work introduces intermediate motion representations to bridge pixels and actions.
Pixel motion fields and optical-flow-like targets provide dense image-plane motion cues for policy learning~\cite{nguyen2025pixel,ranasinghe2025pixel,zheng2025translating}, while point trajectories, 3D point tracks, and object flows introduce stronger geometric or correspondence structure~\cite{bharadhwaj2024track2act,wen2023any,hung20263pointr,huang2026pointworld,dharmarajan2025dream2flow,bharadhwaj2024gen2act}.
Another line uses language-aligned keypoints, value maps, or mark-based prompts to expose task-relevant spatial structure~\cite{huang2024rekep,huang2307composable,liu2024moka,hu2025generalizable}.
These approaches demonstrate that motion and spatial abstractions are often more control-relevant than raw pixels, but they typically use motion as a policy input, an auxiliary prediction target, or a separate planning representation.
JOPAT instead treats 2D tracks with visibility as part of the generative future state itself: the track variables are sampled jointly with visual latents and actions in the same denoising sequence.
This design preserves semantic grounding through visual latents while using tracks as a scalable correspondence-level interface that can be obtained from ordinary videos with modern point trackers~\cite{doersch2023tapvidbenchmarktrackingpoint,karaev2024cotracker3simplerbetterpoint}.

\paragraph{Learning from action-free videos}
Learning from action-free videos is attractive because videos are substantially easier to scale than action-labeled robot demonstrations.
Prior work uses action-free or weakly labeled videos to learn visual representations, dynamics models, inverse dynamics, latent actions, or policy-relevant priors~\cite{pmlr-v162-parisi22a,nair2023r3m,xiao2022masked,seo2022reinforcement,wu2024unleashing,chen2025villa,he2024learning}.
However, generic videos often contain uncontrolled camera motion, unknown agents, and dynamics not caused by the robot, and recent studies suggest that pixel-level pretraining alone does not always translate into improved downstream manipulation success~\cite{li2026matters,shiteleportation,rhoda2026dva}.
JOPAT addresses this issue by using action-free videos to supervise a pixel-track future state rather than actions directly: visual latents preserve object and scene semantics, while tracks and visibility supervise realized correspondences and missing evidence.
Action-labeled robot demonstrations then ground these future-state variables in executable controls.

\section{Preliminaries: world-action models}
\label{sec:problem_wam}

We consider episodic robot manipulation with RGB observations $o_t$ and robot actions $a_t$, where actions are end-effector pose deltas and gripper commands.
At time $t$, the policy receives an observation window $\mathbf{o}^{\mathrm{ctx}}_t=o_{t-h_o+1:t}$ and predicts an executable action chunk $a_{t:t+K-1}$.
An encoder maps this window to a conditioning feature $\mathbf{c}_t=E_c(\mathbf{o}^{\mathrm{ctx}}_t)$.
We assume access to two data sources: a large collection of action-free videos $\mathcal{V}=\{o_{1:T}\}$ and a smaller set of action-labeled demonstrations $\mathcal{D}=\{(o_{1:T},a_{1:T})\}$.
The goal is to learn a policy $\pi_\psi(a_{t:t+K-1}\mid \mathbf{c}_t)$ that transfers scalable motion priors from $\mathcal{V}$ while grounding them in executable robot actions from $\mathcal{D}$.

A world-action model (WAM) provides a generative route to this goal by coupling action prediction with future world-state prediction.
Rather than learning a policy as a direct map from observations to actions, WAM models a joint distribution $p_\theta(a_{t:t+K-1},s_{t:t+H}\mid \mathbf{c}_t)$, where $s_{t:t+H}$ denotes a future state abstraction of the scene.
The policy can then be viewed as the action marginal of this larger predictive model, $\pi_\theta(a_{t:t+K-1}\mid \mathbf{c}_t)=\int p_\theta(a_{t:t+K-1},s_{t:t+H}\mid \mathbf{c}_t)ds$.
This perspective turns policy learning into a question of interface design: \emph{what future world state should a robot predict so that the prediction is useful for action?}
The answer is nontrivial because a future state for control is not the same as a future state for photorealistic reconstruction.
For manipulation, the useful state should expose how robot-controllable scene elements move and whether their evidence remains visible, while still preserving enough visual semantics to identify objects and affordances.

The common instantiation is a pixel-latent WAM.
Let $\mathbf{z}^{o}_{t+H:t+H+1}=E_o(o_{t+H:t+H+1})$ denote compact visual latents of a future observation window and $\mathbf{z}^{a}_{t:t+K-1}$ denote action tokens.
A pixel-latent WAM uses $s^{\mathrm{pix}}_{t:t+H}=\mathbf{z}^{o}_{t+H:t+H+1}$ and learns $p_\theta(\mathbf{z}^{a}_{t:t+K-1},\mathbf{z}^{o}_{t+H:t+H+1}\mid \mathbf{c}_t)$.
Architecturally, this can be implemented by concatenating noisy action and visual tokens and denoising them with a shared transformer,
\begin{equation}
    [\hat{\boldsymbol{\epsilon}}^{a},\hat{\boldsymbol{\epsilon}}^{o}]
    =
    T^{\mathrm{WAM}}_\theta(
    [\tilde{\mathbf{z}}^{a},\tilde{\mathbf{z}}^{o},\mathbf{r}];
    \mathbf{c}_t,\tau_a,\tau_o),
\end{equation}
where $\mathbf{r}$ denotes learnable register tokens and $\tau_a,\tau_o$ are modality-specific diffusion timesteps.
This formulation enables action-free video pretraining through the observation branch, but it also inherits a limitation of pixel prediction: the state interface is dominated by appearance.
For manipulation, appearance is only an indirect carrier of control-relevant change.
A visual latent may encode texture, lighting, background, and camera statistics together with object motion, while the action often depends on sparse correspondence-level facts: which part moved, where it moved, whether it became occluded, and whether it left the visible workspace.
Thus, the central weakness of a pixel-only WAM is not simply insufficient model capacity; it is that the predicted future state does not explicitly expose the motion variables through which actions affect the world.

\section{Methods}
In this section, we first present the overall architecture of the proposed JOPAT, followed by a detailed description of the critical point-track modeling component. We then introduce the training and inference pipeline.
\label{sec:method}

\subsection{Unified pixel-track-action architecture}
\label{subsec:architecture}

\begin{figure*}[t]
    \centering
    \includegraphics[width=\textwidth]{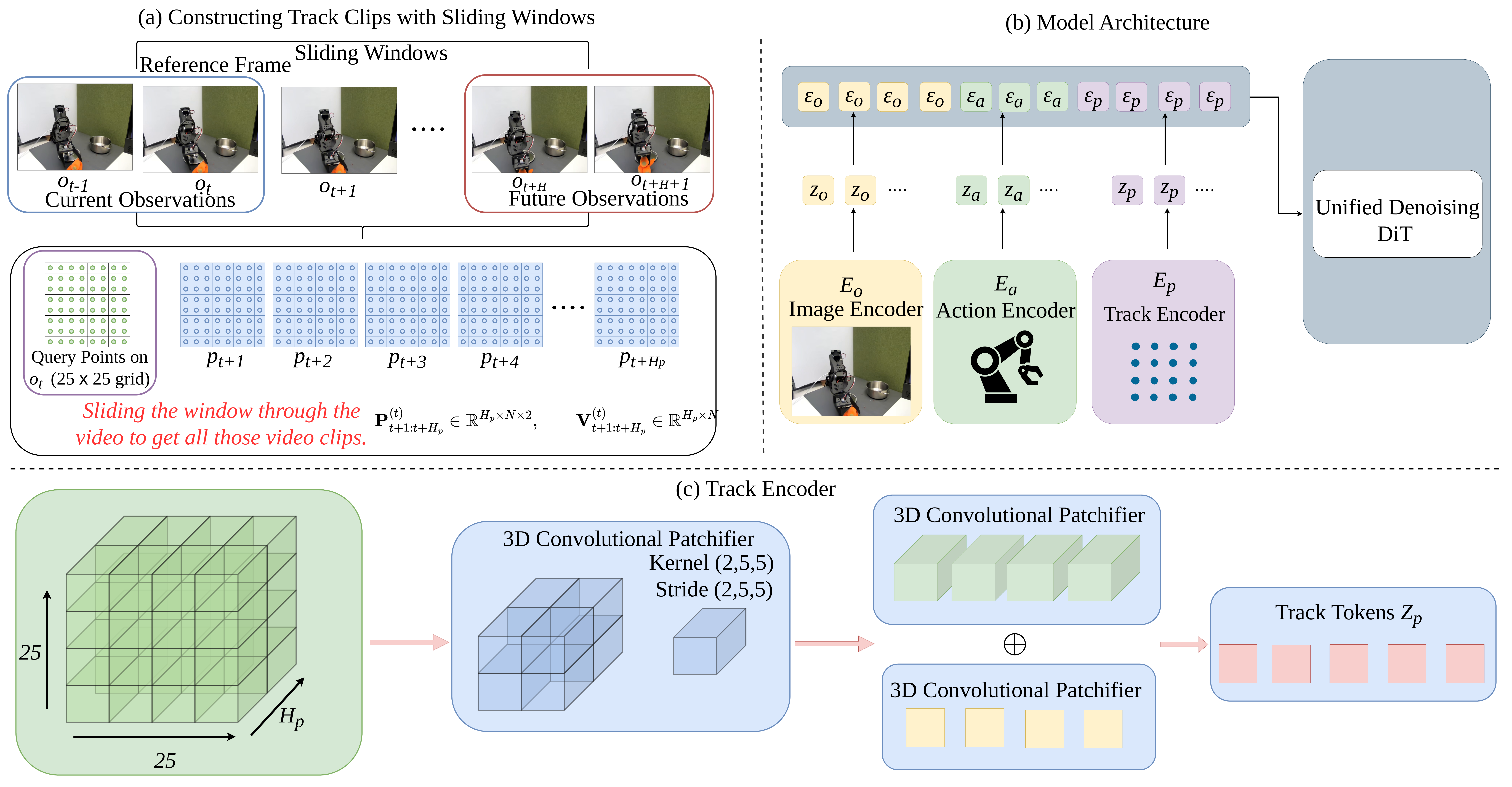}
    \vspace{-0.8em}
    \caption{
    \textbf{Overview of JOPAT.}
    \textbf{(a)} Sliding-window track construction uses the current frame as the reference image for grid query points.
    \textbf{(b)} JOPAT jointly denoises future visual latents, point-track coordinates, and robot actions in a shared Transformer.
    \textbf{(c)} The track-as-video encoder reshapes point tracks into a spatiotemporal grid, applies 3D convolutional patchification, and predicts coordinate noise and visibility logits.
    }
    \vspace{-0.8em}
    \label{fig:jopat_overview}
\end{figure*}

Figure~\ref{fig:jopat_overview} (b) shows the unified JOPAT architecture.
Given the current observation window $o_{t-1:t}$, an observation encoder $E_c$ produces a global conditioning feature $\mathbf{c}_t=E_c(o_{t-1:t})$, which is injected into each DiT block through AdaLN together with modality-specific diffusion timesteps $\tau_a$, $\tau_o$, and $\tau_p$. JOPAT jointly denoises action tokens, future visual-latent tokens, and track tokens in a shared sequence:
\[
    \mathbf{Z}=
    [\tilde{\mathbf{z}}^{a}_{t:t+K-1},
     \tilde{\mathbf{z}}^{o}_{t+H:t+H+1},
     \tilde{\mathbf{z}}^{p}_{t:t+H_p-1},
     \mathbf{r}].
\]
A DiT-style transformer applies bidirectional full attention over this sequence and predicts modality-specific noise:
\[
    \hat{\boldsymbol{\epsilon}}^{a},
    \hat{\boldsymbol{\epsilon}}^{o},
    \hat{\boldsymbol{\epsilon}}^{p}
    =
    T_\theta(\mathbf{Z};\mathbf{c}_t,\tau_a,\tau_o,\tau_p).
\]
Separate heads decode action noise, visual-latent noise, track-coordinate noise, and visibility logits.
This shared denoising design allows action generation to interact directly with imagined future pixels and point tracks, rather than relying on an isolated action decoder with auxiliary prediction heads.
In our implementation, JOPAT conditions on two RGB frames, predicts a two-frame future observation window with offset $H=16$, and uses $H_p=K=19$ for both track and action horizons.

\subsection{Track construction and modeling}
\label{subsec:track_processing}

Figure~\ref{fig:jopat_overview} (a,c) illustrates the track construction and encoding pipeline.
JOPAT conditions on two RGB observations $o_{t-1:t}$ and uses the last frame $o_t$ as the reference frame for point tracking.
For each sliding window, we initialize $N$ grid query points on $o_t$ and track them over $t+1:t+H_p$ with an off-the-shelf point tracker~\cite{karaev2024cotracker3simplerbetterpoint}, producing
\begin{equation}
    \mathbf{P}^{(t)} \in \mathbb{R}^{H_p \times N \times 2},
    \qquad
    \mathbf{V}^{(t)} \in \{0,1\}^{H_p \times N}.
\end{equation}
The superscript $(t)$ denotes that the query points are anchored at the reference frame $o_t$.

To encode tracks compactly, we use a track-as-video representation.
In our implementation, query points form a $25 \times 25$ grid, giving $N=625$.
We reshape the point dimension back to its spatial grid and apply a 3D convolutional patchifier:
\begin{equation}
\begin{aligned}
    \mathbf{P}^{(t)}
    &\rightarrow
    \mathbf{G}^{p} \in \mathbb{R}^{2 \times H_p \times H_g \times W_g},
    \quad H_g W_g=N, \\
    \mathbf{z}^{p}
    &= E_p(\mathbf{P}^{(t)})
     = \mathrm{Patchify}_{3D}(\mathbf{G}^{p}),
    \quad
    \mathbf{z}^{p} \in \mathbb{R}^{L_p \times d}.
\end{aligned}
\end{equation}
After shared denoising, track tokens are decoded into coordinate and visibility predictions:
\begin{equation}
    \hat{\mathbf{P}}^{(t)} = D_p(\hat{\mathbf{z}}^{p}),
    \qquad
    \hat{\mathbf{V}}^{(t)} = D_v(\hat{\mathbf{z}}^{p}),
    \quad
    \hat{\mathbf{V}}^{(t)} \in \mathbb{R}^{H_p \times N}.
\end{equation}
Visibility is predicted only as an output and is not provided to the track encoder, avoiding leakage of ground-truth visibility during denoising.

\subsection{Training and inference}
\label{subsec:training_inference}

JOPAT is trained with a unified denoising objective over action, visual-latent, and track-coordinate tokens.
For each modality $m\in\{a,o,p\}$, let $\mathbf{x}^m_0$ be the clean target and $\boldsymbol{\epsilon}^m\sim\mathcal{N}(0,I)$ be Gaussian noise.
We independently sample a diffusion timestep $\tau_m$ for each modality and construct
\begin{equation}
    \mathbf{x}^m_{\tau_m}
    =
    \sqrt{\bar{\alpha}_{\tau_m}}\mathbf{x}^m_0
    +
    \sqrt{1-\bar{\alpha}_{\tau_m}}\boldsymbol{\epsilon}^m,
    \qquad
    \mathcal{L}_{m}
    =
    \left\|
    \hat{\boldsymbol{\epsilon}}^m
    -
    \boldsymbol{\epsilon}^m
    \right\|_2^2 .
\end{equation}
The track diffusion loss $\mathcal{L}_{p}$ is applied only to 2D coordinates.
Visibility is predicted by a separate head and supervised with binary cross entropy (BCE):
\begin{equation}
    \mathcal{L}_{\mathrm{vis}}
    =
    \frac{1}{H_pN}
    \sum_{\tau=0}^{H_p-1}
    \sum_{i=1}^{N}
    \mathrm{BCE}
    (\hat{\mathbf{V}}_{t+\tau,i},\mathbf{V}_{t+\tau,i}).
\end{equation}
The final objectives for action-labeled demonstrations $\mathcal{D}$ and action-free videos $\mathcal{V}$ are
\begin{equation}
\begin{aligned}
    \mathcal{L}_{\mathcal{D}}
    &=
    \mathcal{L}_{a}
    +
    \lambda_o\mathcal{L}_{o}
    +
    \lambda_p\mathcal{L}_{p}
    +
    \lambda_{\mathrm{vis}}\mathcal{L}_{\mathrm{vis}}, \\
    \mathcal{L}_{\mathcal{V}}
    &=
    \lambda_o\mathcal{L}_{o}
    +
    \lambda_p\mathcal{L}_{p}
    +
    \lambda_{\mathrm{vis}}\mathcal{L}_{\mathrm{vis}}.
\end{aligned}
\end{equation}
Thus, the same architecture learns visual-track dynamics from action-free videos and action-grounded control from robot demonstrations.

At inference time, JOPAT is used for receding-horizon action generation.
Given the current observation context $\mathbf{c}_t$, action, visual, and track tokens are initialized from Gaussian noise and iteratively denoised in the same shared sequence.
Only the action branch is decoded for execution, but the visual and track tokens are not discarded auxiliary outputs: they act as latent future-state variables that interact with action tokens through self-attention during sampling.
Consequently, each action chunk is generated under an implicitly sampled pixel-track future, which is the inference-time counterpart of the track-augmented state interface used during training.
The action decoder outputs a chunk $\hat{a}_{t:t+K-1}$, from which we execute the first action or a short prefix before replanning.

\paragraph{Implementation details}
We use a DiT-style Transformer with depth $12$, hidden dimension $768$, and $12$ attention heads.
The model uses $8$ learnable register tokens.
For tracks, we use $N=625$ query points arranged as a $25\times25$ grid and a spatiotemporal patch size of $(2,5,5)$.
We train with a DDIM scheduler using $100$ diffusion steps and sample with $10$ denoising steps at inference.

\section{Experiments}
\label{sec:experiments}

\paragraph{Experiment design} Our experiments are organized to examine the state-interface hypothesis introduced in~\cref{sec:Intro}.
(i) a track-augmented future state should improve manipulation policy learning over pixel-centric world-action modeling because it exposes controllable correspondences that pixels leave implicit.
(ii) the gains should come from the complementary roles of visual latents, tracks, and visibility: semantics, motion correspondences, and missing-evidence state.
(iii) action-free video pretraining should be useful when it supervises the same pixel-track future state that later mediates action generation under limited robot demonstrations.

We therefore evaluate JOPAT along four axes.
\textbf{(Q1) Main performance:} Does JOPAT improve success rates over prior policy-learning and world-action baselines on standard manipulation benchmarks?
\textbf{(Q2) Representation design:} Are visual latents, point tracks, and visibility all necessary for this future-state interface under partial observability?
\textbf{(Q3) Robustness:} Does the pixel-track future state remain reliable under long-horizon distribution shift, occlusion, and distractors?
\textbf{(Q4) Scalable pretraining:} Does action-free video pretraining help under limited demonstrations, and does the benefit extend beyond robot-domain videos?

We answer Q1 on LIBERO~\cite{liu2023libero}, Q2 through modality and visibility ablations, Q3 with a modified LIBERO-Long out-of-distribution (OOD) stress test and real-robot LeRobot SO-101 experiments, and Q4 with action-free pretraining studies using robot-domain DROID videos~\cite{khazatsky2024droid} and generic OpenVid-1M videos~\cite{nan2024openvid1m}.
For simulation, we evaluate all four LIBERO suites and use LIBERO-90 for the standard pretraining protocol before target-suite finetuning.
For pretraining ablations, we compare no video pretraining, DROID action-free pretraining, and a matched 10k-video subset from OpenVid-1M; all video-only pretraining masks the action branch and supervises only visual latents, point tracks, and visibility.

\begin{table*}[t]
\centering
\caption{\textbf{Simulation experimental results.}
Comparison of task success rates (SR) and ranks (RK) on the LIBERO benchmark across four task types.
``\dag'' indicates our reproduced results.}
\label{tab:libero-p}
\small
\setlength{\tabcolsep}{1.6pt}

\newcolumntype{Z}{>{\hsize=2.8\hsize\raggedright\arraybackslash}X}
\newcolumntype{Y}{>{\hsize=0.82\hsize\centering\arraybackslash}X}
\begin{tabularx}{\textwidth}{Z|YY|YY|YY|YY|YY}
\toprule
\textbf{Method}
& \multicolumn{2}{c|}{\textbf{Spatial}}
& \multicolumn{2}{c|}{\textbf{Object}}
& \multicolumn{2}{c|}{\textbf{Goal}}
& \multicolumn{2}{c|}{\textbf{Long}}
& \multicolumn{2}{c}{\textbf{Average}} \\
& SR $\uparrow$ & RK $\downarrow$
& SR $\uparrow$ & RK $\downarrow$
& SR $\uparrow$ & RK $\downarrow$
& SR $\uparrow$ & RK $\downarrow$
& SR $\uparrow$ & RK $\downarrow$ \\
\midrule
Diffusion Policy~\cite{chi2024diffusionpolicyvisuomotorpolicy}
& 78.3 & 15 & 92.5 & 10 & 68.3 & 15 & 50.5 & 15 & 72.4 & 15 \\
Octo fine-tuned~\cite{octomodelteam2024octoopensourcegeneralistrobot}
& 78.9 & 14 & 85.7 & 15 & 84.6 & 12 & 51.1 & 14 & 75.1 & 14 \\
OpenVLA~\cite{kim2024openvlaopensourcevisionlanguageactionmodel}
& 84.7 & 11 & 88.4 & 14 & 79.2 & 13 & 53.7 & 13 & 76.5 & 13 \\
$\pi_0$ fine-tuned~\cite{black2026pi0visionlanguageactionflowmodel}
& 96.8 & 5 & 98.8 & 2 & 95.8 & 4 & 85.2 & 7 & 94.2 & 7 \\
$\pi_0$-Fast~\cite{pertsch2025fastefficientactiontokenization}
& 96.4 & 6 & 96.8 & 7 & 88.6 & 8 & 60.2 & 11 & 85.5 & 8 \\
$\pi_{0.5}$-KI~\cite{intelligence2025pi05visionlanguageactionmodelopenworld}
& 98.0 & 2 & 97.8 & 6 & 95.6 & 5 & 85.8 & 6 & 96.0 & 4 \\
OpenVLA-OFT~\cite{kim2025finetuningvisionlanguageactionmodelsoptimizing}
& 97.6 & 3 & 98.4 & 4 & 97.9 & 2 & 94.5 & 3 & 97.1 & 3 \\
SpatialVLA~\cite{qu2025spatialvlaexploringspatialrepresentations}
& 88.2 & 9 & 89.9 & 13 & 78.6 & 14 & 55.5 & 12 & 78.1 & 12 \\
PD-VLA\dag~\cite{song2025acceleratingvisionlanguageactionmodelintegrated}
& 95.5 & 7 & 96.7 & 8 & 94.9 & 7 & 91.7 & 4 & 94.7 & 5 \\
STAR~\cite{li2025starlearningdiverserobot}
& 95.5 & 7 & 98.3 & 5 & 95.0 & 6 & 88.5 & 5 & 94.3 & 6 \\
Dita~\cite{hou2025ditascalingdiffusiontransformer}
& 84.2 & 12 & 96.3 & 9 & 85.4 & 11 & 63.8 & 10 & 82.4 & 11 \\
CoT-VLA~\cite{zhao2025cotvlavisualchainofthoughtreasoning}
& 87.5 & 10 & 91.6 & 12 & 87.6 & 9 & 69.0 & 9 & 83.9 & 10 \\
CogVLA~\cite{li2025cogvlacognitionalignedvisionlanguageactionmodel}
& \textbf{98.6} & \textbf{1} & 98.8 & 2 & 96.6 & 3 & 95.4 & 2 & 97.4 & 2 \\
\midrule
UWM~\cite{zhu2025unifiedworldmodelscoupling}
& 82.3 & 13 & 92.2 & 11 & 86.8 & 10 & 77.6 & 8 & 84.7 & 9 \\
\rowcolor[HTML]{ECDFF2}
JOPAT
& 97.2 & 4 & \textbf{98.9} & \textbf{1} & \textbf{98.4} & \textbf{1} & \textbf{96.4} & \textbf{1} & \textbf{97.8} & \textbf{1} \\
\bottomrule
\end{tabularx}
\vspace{-1em}
\end{table*}

\subsection{Simulation results}
\label{subsec:libero}

We evaluate JOPAT on the standard LIBERO benchmark, which contains four suites:
\textit{Spatial}, \textit{Object}, \textit{Goal}, and \textit{Long}, with 10 tasks and 50 demonstrations per task in each suite.
Following the benchmark protocol, we report the average success rate (SR) over tasks within each suite.
JOPAT is pretrained on LIBERO-90 for 100K steps and finetuned on each target suite for 10K steps.

Table~\ref{tab:libero-p} reports the main simulation results.
JOPAT achieves the best average performance across the four suites and obtains the largest improvement on \textit{Long}, the most temporally extended suite.
This indicates that joint pixel-and-track prediction is particularly beneficial when policy learning requires maintaining action-relevant state over longer horizons. The gain on \textit{Long} is consistent with the motivation of JOPAT.
Compared with pixel-centric prediction, point tracks provide a compact motion-centric interface, while visual latents retain semantic and contextual information.
We further isolate these components in the ablation study in Section~\ref{subsec:joint_ablation}.

\subsection{Real-world experiments}
\label{sec:real-robot}

We further evaluate real-world transfer on a physical LeRobot SO-101 platform.
The benchmark includes four tasks spanning complementary challenges:
\textit{Cook-Soup} for long-horizon sequencing under repeated interactions,
\textit{Insert-Peg} for precision alignment and contact-sensitive execution,
\textit{Push-Tomato} for contact-rich object displacement under pose variation,
and \textit{Pick-Grocery} for object-level OOD generalization.
All real-robot training, inference, evaluation, and task setup details are provided in Appendix~\ref{app:real_robot_protocol}.

\begin{figure*}[ht]
    \centering
    \setlength{\figurewidth}{0.235\textwidth}
    \input{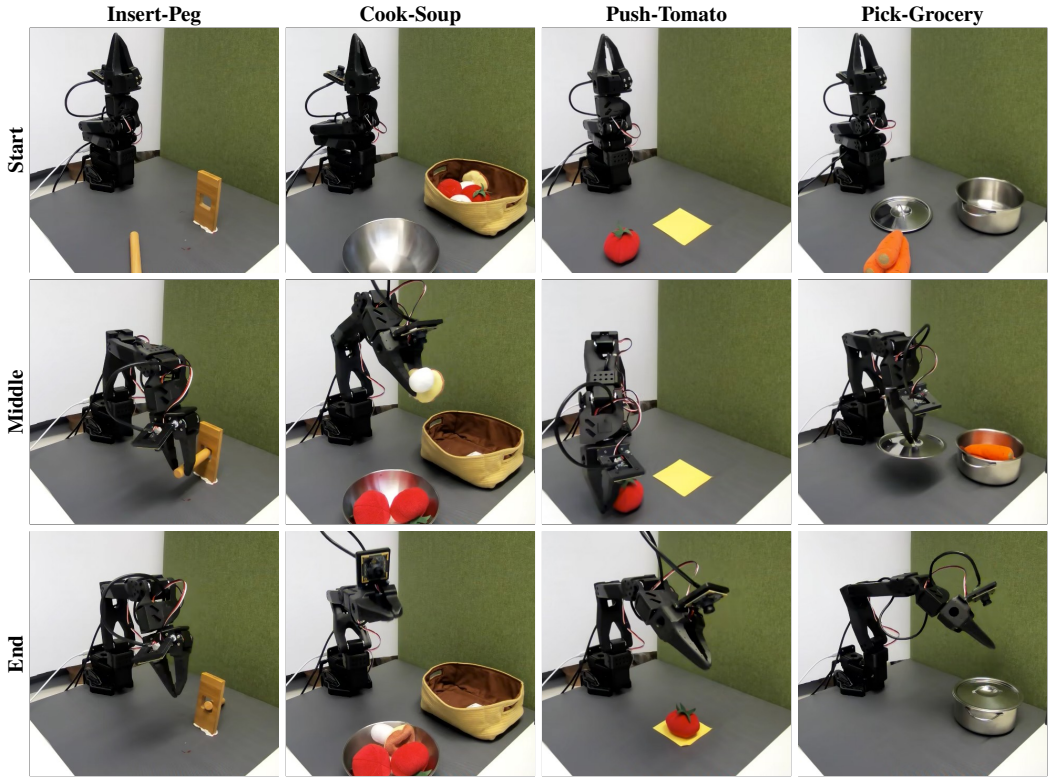}
    \caption{\textbf{Real-robot task setup.}
    Insert-Peg, Cook-Soup, Push-Tomato, and Pick-Grocery on the LeRobot SO-101 platform.
    The first row shows the initial configuration, the second row shows an intermediate state, and the third row shows successful task completion.}
    \label{fig:lerobot_setup}
    \vspace{-1em}
\end{figure*}

\begin{table}[ht]
\centering
\caption{\textbf{Real-robot performance on LeRobot.}
SR on four real-world tasks are reported.}
\label{tab:lerobot}
\small
\setlength{\tabcolsep}{4pt}
\renewcommand{\arraystretch}{1.02}
\begin{tabular}{lccccc}
\toprule
\textbf{Method}
& \textbf{Cook-Soup}
& \textbf{Insert-Peg}
& \textbf{Push-Tomato}
& \textbf{Pick-Grocery}
& \textbf{Avg.} \\
\midrule
ACT~\cite{zhao2023learningfinegrainedbimanualmanipulation}
& 40\% & 0\% & 70\% & 50\% & 40\% \\
UWM~\cite{zhu2025unifiedworldmodelscoupling}
& 10\% & 0\% & 80\% & 40\% & 32.5\% \\
\midrule
\rowcolor[HTML]{ECDFF2}
JOPAT (Ours)
& \textbf{60\%} & \textbf{10\%} & \textbf{100\%} & \textbf{60\%} & \textbf{57.5\%} \\
\bottomrule
\end{tabular}
\vspace{-2mm}
\end{table}

Table~\ref{tab:lerobot} shows that JOPAT achieves the best average success rate, outperforming ACT and UWM by 17.5 and 25.0 points, respectively.
The largest gain appears on \textit{Push-Tomato}, where JOPAT reaches 100\% SR, suggesting that explicit track prediction provides a useful motion interface for contact-rich object displacement.
JOPAT also improves on \textit{Cook-Soup}, consistent with its robustness to occlusion and long-horizon state drift observed in simulation.
By contrast, \textit{Insert-Peg} remains difficult for all methods, indicating that millimeter-level insertion may require additional geometric calibration or contact-aware sensing beyond RGB-based track-and-latent modeling.

\subsection{Ablation on joint pixel-track modeling}
\label{subsec:joint_ablation}

We study whether the above performance gains come from joint pixel-and-track modeling and create three ablated variants:
\textbf{Latent-only}, which removes track prediction;
\textbf{Track-only}, which removes visual-latent prediction;
and \textbf{Joint}, which predicts visual latents, tracks, visibility, and actions with a shared denoising transformer.
All variants use the same pretraining data, optimization schedule, and downstream finetuning protocol. 

\paragraph{Simulation ablation} The results in Table~\ref{tab:ablation_multimodality} show that joint modeling substantially outperforms both single-modality variants, improving over Latent-only by 19.0 points and over Track-only by 70.2 points in average SR.
This supports our hypothesis: track tokens provide correspondence-level motion constraints for action-token denoising, while visual latents preserve the semantic context needed to decide which object, affordance, or task state matters.
Latent-only removes these motion constraints, leading to failures when correct object recognition is not enough to maintain precise long-horizon motion; Track-only removes visual grounding, leading to failures when similar motions must be tied to the correct object, affordance, or task state.
The joint model succeeds by coupling both signals in the same denoising state.\

\begin{table*}[ht]
\centering
\caption{\textbf{Joint vs. single-modality ablation on LIBERO-Long.}
Success rates (SR, \%) across ten long-horizon tasks. All variants are pretrained on DROID action-free videos under identical settings.}
\label{tab:ablation_multimodality}
\footnotesize
\setlength{\tabcolsep}{4.5pt}
\renewcommand{\arraystretch}{1.05}
\begin{tabular*}{\textwidth}{@{\extracolsep{\fill}}lccc@{\qquad}lccc@{}}
\toprule
\textbf{Task} & \textbf{Latent-only} & \textbf{Track-only} & \textbf{Joint}
& \textbf{Task} & \textbf{Latent-only} & \textbf{Track-only} & \textbf{Joint} \\
\midrule
Mug-Mug           & 58 &  2 & \textbf{92}  & Tomato-Sauce      & 93 & 48 & \textbf{100} \\
Soup-Cheese       & 78 & 14 & \textbf{94}  & Cream-Butter      & 91 & 66 & \textbf{100} \\
Moka-Moka         & 67 & 26 & \textbf{89}  & Microwave         & 94 & 32 & \textbf{100} \\
Book-Candy        & 63 & 14 & \textbf{100} & Chocolate-Pudding & 73 & 46 & \textbf{91}  \\
Bowl-Drawer       & 95 &  2 & \textbf{100} & Moka-Stove        & 62 & 12 & \textbf{98}  \\
\midrule
\rowcolor[HTML]{F4ECF7}
\multicolumn{8}{c}{\textbf{Average SR:} Latent-only 77.4 \quad Track-only 26.2 \quad Joint \textbf{96.4}} \\
\bottomrule
\vspace{-1em}
\end{tabular*}
\end{table*}

\paragraph{Real-robot ablation} The similar effect appears in real robot experiments in Table~\ref{tab:real_modality_ablation}.
Joint modeling improves average SR from 35.0\% for Latent-only and 15.0\% for Track-only to 57.5\%, corresponding to gains of 22.5 and 42.5 points.
Latent-only retains object appearance but leaves contact-relevant motion implicit, while Track-only exposes motion structure but lacks the visual grounding needed to identify the correct object and goal region.
JOPAT improves every task under the same data budget, although Insert-Peg remains difficult for all variants, suggesting that millimeter-level insertion may require additional geometry or contact information beyond RGB latents and 2D tracks.

\begin{table}[ht]
\centering
\caption{\textbf{Real-robot modality ablation.} SR are reported on four LeRobot SO-101 tasks.}
\label{tab:real_modality_ablation}
\footnotesize
\setlength{\tabcolsep}{4pt}
\begin{tabular}{lccccc}
\toprule
\textbf{Method} & \textbf{Cook-Soup} & \textbf{Insert-Peg} & \textbf{Push-Tomato} & \textbf{Pick-Grocery} & \textbf{Avg.} \\
\midrule
Latent-only                                  & 40          & 0           & 70           & 30          & 35.0 \\
Track-only                                   & 10          & 0           & 30           & 20          & 15.0 \\
\rowcolor[HTML]{ECDFF2} Joint (Ours)         & \textbf{60} & \textbf{10} & \textbf{100} & \textbf{60} & \textbf{57.5} \\
\bottomrule
\vspace{-1em}
\end{tabular}
\end{table}

\subsection{Ablation on track visibility prediction}

We further ablate visibility prediction in Table~\ref{tab:ablation_visibility}.
Removing the visibility head reduces real-robot average SR from 57.5\% to 47.5\%, with the largest improvement from visibility on Cook-Soup.
This supports the role of visibility supervision as a missing-evidence constraint on the track stream: the model learns when a correspondence is unobservable instead of forcing every point to explain a visible coordinate.
The unchanged Pick-Grocery result indicates that visibility is not merely extra capacity; it mainly helps tasks with self-occlusion, transient out-of-view motion, or long-horizon object motion where confusing hidden points with stationary points can cause drift.
\begin{table}[ht]
\centering
\caption{\textbf{Track visibility ablation.} We report the SR on LeRobot SO-101 tasks. The visibility head improves robustness under self-occlusion and transient out-of-view motion.}
\label{tab:ablation_visibility}
\setlength{\tabcolsep}{4pt}
\begin{tabular}{lccccc}
\toprule
Method & Cook-Soup & Insert-Peg & Push-Tomato & Pick-Grocery & Avg. \\
\midrule
w/o visibility                              & 40          & 0           & 90           & 60          & 47.5 \\
\rowcolor[HTML]{ECDFF2} w/ visibility (Ours) & \textbf{60} & \textbf{10} & \textbf{100} & \textbf{60} & \textbf{57.5} \\
\bottomrule
\vspace{-3em}
\end{tabular}
\end{table}

\subsection{Does action-free video pretraining help?}
\label{subsec:pretraining_scaling}

Table~\ref{tab:scaling} isolates the effect of action-free video pretraining on LIBERO-Long.
Both pretraining sources consistently outperform training from scratch across all demonstration budgets, showing that JOPAT's pixel-track future-state objective turns action-free videos into useful motion priors for downstream control.
The gain is especially large in the low-data regime: with only 10 demonstrations, DROID pretraining improves SR from 11.9\% to 64.2\%, and even generic OpenVid-1M pretraining reaches 48.5\%.
This indicates that the benefit is not merely due to robot-domain proximity; the model can also extract transferable object-motion and occlusion priors from non-robot videos.
At 50 demonstrations, the gap between DROID and OpenVid-1M nearly vanishes (96.4\% vs. 95.1\%), suggesting that domain-specific pretraining is most valuable when supervision is scarce, while generic video priors can be aligned effectively once sufficient action-labeled demonstrations are available.

\begin{table}[ht]
\centering
\caption{\textbf{Action-free video pretraining on LIBERO-Long.} SR are reported.}
\label{tab:scaling}
\footnotesize
\begin{tabular}{lccc}
\toprule
\textbf{Method} & \textbf{10 demos} & \textbf{25 demos} & \textbf{50 demos} \\
\midrule
JOPAT w/o PT          & 11.9\%          & 31.6\%          & 66.1\% \\
JOPAT (DROID PT)      & \textbf{64.2}\% & 82.7\%          & \textbf{96.4}\% \\
JOPAT (OpenVid-1M PT) & 48.5\%          & \textbf{84.6}\% & 95.1\% \\
\bottomrule
\end{tabular}
\vspace{-3mm}
\end{table}

\subsection{Robustness on OOD tasks}
\label{subsec:ood}

To evaluate robustness under stronger partial observability and distribution shift, we construct a modified LIBERO-Long setting with three changes: 
(1) the object initialization range is expanded by 5 cm;
(2) one background object is replaced with an unseen object in each task;
and (3) an unseen distractor is placed in front of the target object.
All modifications are applied consistently across all compared methods.
This setting increases failures caused by target-object occlusion, distractor interference, larger object displacement, and delayed errors in long-horizon execution.

\begin{table*}[ht]
\caption{\textbf{OOD evaluation on modified LIBERO-Long.}
SR are reported.}
\label{tab:ood_lerobot}
\centering
\begin{tabular}{l ccc >{\columncolor[HTML]{ECDFF2}}c}
\toprule
Task & Diffusion Policy~\cite{chi2024diffusionpolicyvisuomotorpolicy} & $\pi_{0.5}$ fine-tuned~\cite{intelligence2025pi05visionlanguageactionmodelopenworld} & UWM~\cite{zhu2025unifiedworldmodelscoupling} & JOPAT (Ours) \\
\midrule
Book-Candy   & 0.24 & \textbf{0.95} & 0.38 & 0.88 \\
Soup-Cheese  & 0.21 & \textbf{0.82} & 0.44 & 0.78 \\
Bowl-Drawer  & 0.56 & 0.12 & 0.52 & \textbf{0.68} \\
Moka-Moka    & 0.44 & \textbf{0.70} & 0.22 & 0.48 \\
Mug-Mug      & 0.18 & 0.08 & 0.14 & \textbf{0.50} \\
\midrule
Avg.         & 0.32 & 0.53 & 0.34 & \textbf{0.66} \\
\bottomrule
\end{tabular}
\end{table*}

Table~\ref{tab:ood_lerobot} shows that JOPAT achieves the best average SR, improving over Diffusion Policy, UWM, and fine-tuned $\pi_{0.5}$ by 34, 32, and 13 points, respectively.
Compared with UWM, JOPAT improves on all five tasks, with especially large gains on \textit{Mug-Mug} (+36 points), \textit{Moka-Moka} (+26 points), and \textit{Book-Candy} (+50 points).
These tasks require maintaining object state across extended interactions while avoiding distractor-induced drift, suggesting that explicit correspondence and visibility prediction provides a stable future-state interface under long-horizon distribution shift.

Interestingly, fine-tuned $\pi_{0.5}$ achieves the best performance on \textit{Book-Candy}, \textit{Soup-Cheese}, and \textit{Moka-Moka}, but performs poorly on \textit{Bowl-Drawer} and \textit{Mug-Mug}.
This leads to a lower average SR than JOPAT despite strong peak performance.
The result suggests that large pretrained VLA models can be highly effective on some OOD tasks, but may be less stable across task families when distractors and occlusions alter the visual grounding conditions.
In contrast, JOPAT is not always the best on every individual task, but achieves the strongest average performance and the best results on tasks where distractor-aware object tracking is critical.



\section{Conclusion and discussion}
\label{sec:conclusion}
In this work, we introduced JOPAT, a unified world-action model that treats 2D point tracks and visual latents as joint predictive modalities for action generation. By making motion correspondences explicit alongside pixel-level visual latents, our approach anchors predictions on geometric structures and improves robustness against occlusion. Experiments on LIBERO and real-world LeRobot tasks confirm that this formulation outperforms pixel-based baselines in data efficiency, effectively transferring motion priors from action-free video to downstream control.

\paragraph{Limitations} Despite these results, limitations persist. First, the sparsity of grid-based point tracks may overlook fine-grained deformations required for sub-centimeter dexterity. Second, our reliance on off-the-shelf trackers (e.g., CoTracker) creates a performance ceiling; the model cannot learn dynamics that the supervisor fails to capture. Additionally, while efficient, the unified transformer’s inference speed remains a bottleneck for high-frequency control (>20Hz), and the current architecture is optimized for static cameras, limiting applicability to mobile manipulation scenarios where ego-motion complicates tracking.
\section*{Acknowledgments}

AS and JK acknowledge funding from the Research Council of Finland (352788, 362407, 373999, 373780, 362408, and 339730). This work was supported by the Research Council of Finland Flagship programme: Finnish Center for Artificial Intelligence FCAI. We acknowledge the computational resources provided by the Aalto Science-IT project.

\bibliographystyle{unsrtnat}
\bibliography{references}

\clearpage
\appendix
\section{Supplementary material}
\label{app:supplement}

This appendix provides implementation details and supplementary analyses that complement the main paper. We include the real-robot experimental protocol, the real-robot action-free pretraining result, prediction-horizon sensitivity, and task-level failure modes.

\subsection{Implementation details}
\label{app:implementation}

\paragraph{Architecture}
JOPAT uses a DiT-style transformer backbone with modality-specific encoders and decoders for visual latents, point tracks, and actions. The conditioning encoder processes the two most recent RGB observations and injects the resulting global feature into each Transformer block through AdaLN. Future observations are encoded with a frozen SDXL VAE, while point tracks are represented on a regular query grid and patchified with a 3D convolutional encoder. The Transformer output is split into modality-specific heads for action noise, visual-latent noise, track-coordinate noise, and visibility logits, matching the unified pixel-track-action architecture in Section~\ref{subsec:architecture} and the track construction in Section~\ref{subsec:track_processing}.

\paragraph{Training data}
For action-free videos, the action branch is masked and the model is trained only with visual-latent, track-coordinate, and visibility losses. For action-labeled robot demonstrations, all branches are supervised. Point-track and visibility targets are extracted with CoTrackerV3~\cite{karaev2024cotracker3simplerbetterpoint} using the sliding-window construction described in Section~\ref{subsec:track_processing}. Simulation and real-robot experiments use the same joint denoising formulation, with dataset-specific pretraining and finetuning protocols described in the corresponding experiment sections.

\paragraph{Inference}
At test time, JOPAT performs receding-horizon action generation with DDIM sampling. Action, visual, and track tokens are initialized from Gaussian noise and denoised conditioned on the current observation history. The model can also sample future visual latents and tracks for analysis, but policy execution uses only the decoded action chunk; the first 8 actions are executed before replanning.

\paragraph{Compute resources.}
Action-free pretraining was performed on 4 NVIDIA H200 GPUs for approximately 5 days. Task-specific finetuning used 1 NVIDIA H200 GPU for approximately 1 day. During deployment, inference runs on a single NVIDIA RTX 4090 GPU at approximately 10 Hz.

\begin{table*}[t]
    \centering
    \caption{\textbf{JOPAT hyperparameters.} Unless otherwise stated, the same configuration is used for simulation and real-robot experiments.}
    \label{tab:hyperparams}
    \vspace{-1mm}
    \footnotesize
    \setlength{\tabcolsep}{6pt}
    \renewcommand{\arraystretch}{1.05}
    \begin{tabular}{@{}ll@{}}
        \toprule
        \textbf{Parameter} & \textbf{Value} \\
        \midrule
        \multicolumn{2}{l}{\textbf{Data and representation}} \\
        Observation history & 2 RGB frames \\
        Image resolution & $224 \times 224$ \\
        Future-observation offset & $H=16$ steps \\
        Action / track horizon & $K=H_p=19$ steps \\
        Executed action prefix & 8 steps \\
        SDXL latent shape & $28 \times 28 \times 4$ \\
        Query-point grid & $25 \times 25$ ($N=625$) \\
        Track patch size & $(2,5,5)$ over time and grid dimensions \\
        \midrule
        \multicolumn{2}{l}{\textbf{Model}} \\
        Conditioning encoder & ResNet-18, ImageNet initialization \\
        Visual target encoder & Frozen SDXL VAE~\cite{podell2023sdxlimprovinglatentdiffusion} \\
        Transformer depth & 12 layers \\
        Hidden dimension & 768 \\
        Attention heads & 12 \\
        MLP ratio & 4 \\
        Register tokens & 8 \\
        Conditioning mechanism & AdaLN \\
        \midrule
        \multicolumn{2}{l}{\textbf{Diffusion and optimization}} \\
        Noise schedule & squaredcos\_cap\_v2 \\
        Training diffusion steps & 100 \\
        Inference denoising steps & 10 \\
        Sampler & DDIM~\cite{song2022denoisingdiffusionimplicitmodels} \\
        Optimizer & AdamW~\cite{loshchilov2019decoupledweightdecayregularization} \\
        Learning rate & $1\times10^{-4}$ \\
        Weight decay & $1\times10^{-6}$ \\
        Adam betas & $(0.9,0.999)$ \\
        Batch size & 144 for pretraining; 72 for finetuning \\
        LR schedule & Constant for pretraining; cosine for finetuning \\
        Warmup steps & 1000 \\
        Loss weights & $\lambda_a=\lambda_o=\lambda_p=\lambda_{\mathrm{vis}}=1$ \\
        \bottomrule
    \end{tabular}
    \vspace{-2mm}
\end{table*}

\subsection{Real-robot experimental protocol}
\label{app:real_robot_protocol}

\paragraph{Platform and tasks}
We evaluate on the LeRobot SO-101 platform with four manipulation tasks:
\textit{Cook-Soup}, which requires long-horizon sequencing and repeated object interactions;
\textit{Insert-Peg}, which requires precise geometric alignment and contact-sensitive execution;
\textit{Push-Tomato}, which tests contact-rich object displacement under pose and interaction variation;
and \textit{Pick-Grocery}, which tests object-level OOD generalization to unseen objects.
Together, these tasks evaluate robustness to partial observability, arm-object occlusion, contact-induced dynamics, and distribution shift.

\paragraph{Training data and finetuning}
For real-robot transfer, JOPAT is first pretrained on DROID action-free videos~\cite{khazatsky2024droid}.
During action-free pretraining, the action branch is masked and only the visual-latent, track-coordinate, and visibility objectives are applied.
Point tracks and visibility labels are extracted with CoTrackerV3~\cite{karaev2024cotracker3simplerbetterpoint} using the same sliding-window construction as in the simulation experiments.
The pretrained model is then finetuned separately on each LeRobot task using 50 action-labeled demonstrations.
During finetuning, all branches are supervised so that the pretrained pixel-track motion priors are aligned with executable SO-101 actions.

\paragraph{Inference and evaluation}
At test time, JOPAT uses receding-horizon action generation with DDIM sampling.
The policy conditions on the two most recent RGB observations, denoises action, visual, and track tokens jointly, and executes only the decoded action chunk.
Unless otherwise stated, we use a future-observation offset of $H=16$, an action/track horizon of $19$ steps, 10 DDIM denoising steps, and execute the first 8 actions before replanning.
We report success rate over 10 evaluation rollouts per task for each method.

\paragraph{Task setup}
Figure~\ref{fig:lerobot_setup} visualizes the four LeRobot SO-101 tasks.
Each column corresponds to one task, and the three rows show the initial configuration, an intermediate state, and a successful final state.

\subsection{Additional ablations}
\label{app:ablations}

The main paper reports the LIBERO-Long joint-versus-single-modality ablation in Table~\ref{tab:ablation_multimodality}, the real-robot modality ablation in Table~\ref{tab:real_modality_ablation}, the visibility ablation in Table~\ref{tab:ablation_visibility}, and the LIBERO-Long pretraining scaling study in Table~\ref{tab:scaling}. Here we provide two supplementary real-robot analyses: the effect of DROID action-free video pretraining and the sensitivity to the future-observation offset. Both follow the real-robot protocol in Appendix~\ref{app:real_robot_protocol}.

\subsubsection{Action-free pretraining}
\label{app:pretraining_ablation}

\begin{figure}[t]
\centering
\includegraphics[width=\linewidth]{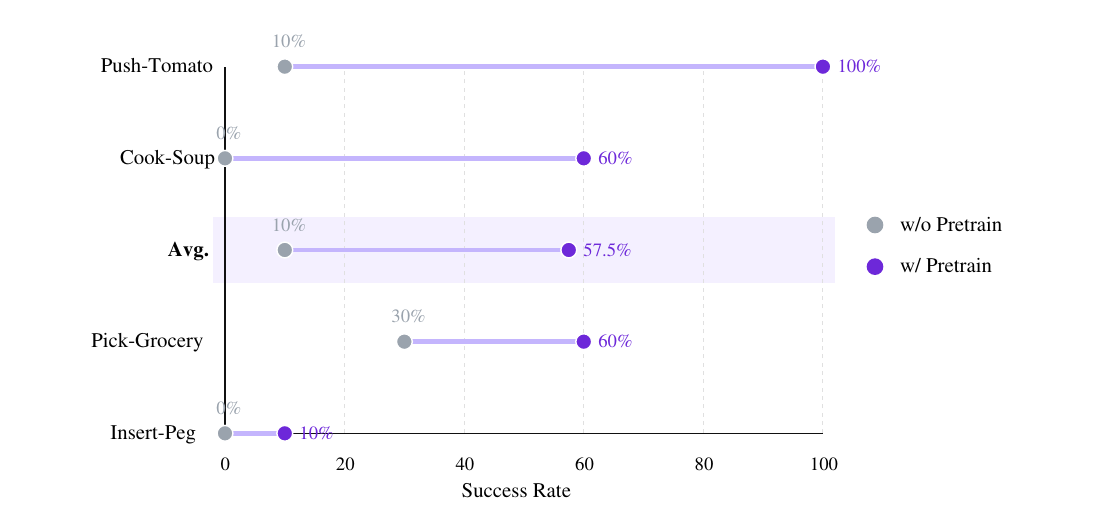}
\caption{\textbf{Action-free pretraining ablation.} Average real-robot success rate with and without DROID action-free video pretraining.}
\label{fig:app_pretraining_ablation}
\vspace{-3mm}
\end{figure}

Figure~\ref{fig:app_pretraining_ablation} summarizes the effect of action-free video pretraining. With the same 50 real-robot finetuning demonstrations per task, DROID pretraining improves the average success rate from $10.0\%$ to $57.5\%$. The largest gains occur on Push-Tomato and Cook-Soup, suggesting that action-free videos provide useful priors for object motion, occlusion handling, and contact-induced dynamics. Insert-Peg remains difficult, indicating that high-precision insertion also requires fine-grained 3D geometry or contact sensing beyond 2D tracks and RGB latents.

\subsubsection{Prediction-horizon sensitivity}
\label{app:horizon_sensitivity}

\begin{figure}[t]
\centering
\hspace*{9mm}%
\includegraphics[
  width=\linewidth,
  trim=0 0 9mm 0,
  clip
]{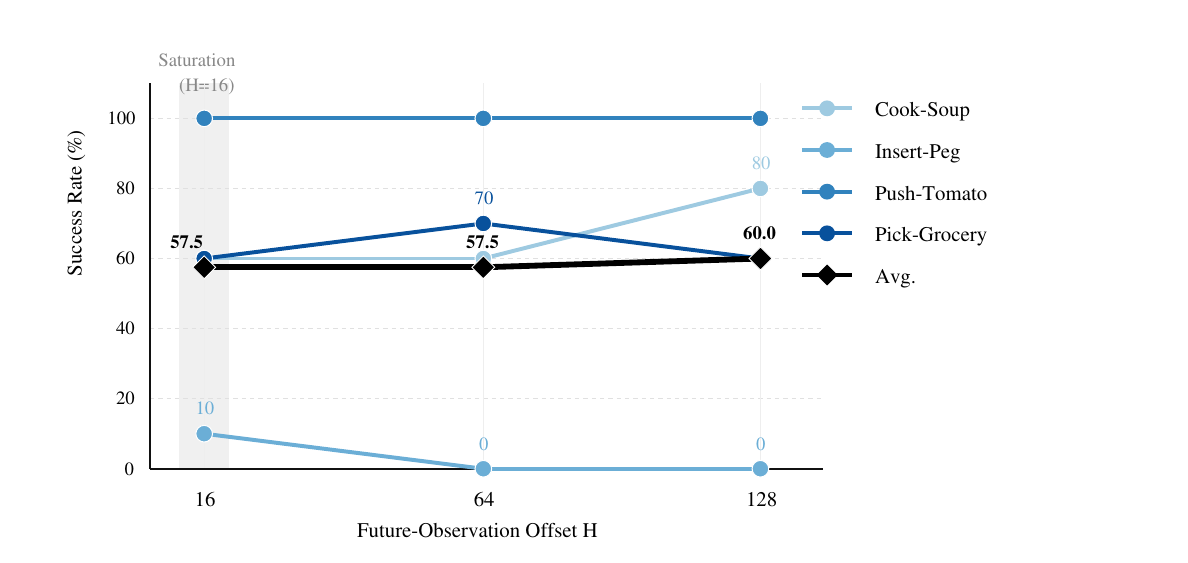}
\caption{\textbf{Horizon sensitivity.} Average real-robot success rate for different future-observation offsets.}
\label{fig:app_horizon_sensitivity}
\vspace{-3mm}
\end{figure}

Figure~\ref{fig:app_horizon_sensitivity} shows that moderate future-observation offsets already capture the dynamics needed for most tasks. The $H=16$ setting matches the performance of longer offsets on average, while avoiding the higher uncertainty associated with very long future-track prediction. Longer offsets can help multi-stage tasks such as Cook-Soup, but may hurt precision-sensitive tasks such as Insert-Peg, where future uncertainty produces noisy supervision for fine contact alignment.

\subsection{Qualitative behavior and failure modes}
\label{app:failure_modes}

\paragraph{Occlusion and off-screen motion}
In qualitative rollouts, the predicted tracks often remain temporally coherent when the robot arm occludes the target or when objects move near the image boundary. This behavior is consistent with the quantitative visibility ablation: the model learns to maintain a belief over hidden object motion rather than relying only on currently visible pixels. Such object permanence is particularly useful for long-horizon tasks in which the policy must continue acting through temporary perceptual dropout.

\paragraph{Task-level failure modes}
The remaining failures are task dependent:
\begin{itemize}[leftmargin=*]
    \item \textbf{Cook-Soup.} Failures often occur around the pot lid, where specular reflections and unusual lid orientations weaken visual grounding. Stronger illumination augmentation or depth cues may improve robustness.
    \item \textbf{Insert-Peg.} The task requires millimeter-level 3D alignment between the peg and socket. Two-dimensional tracks reduce perceptual drift but do not fully resolve contact geometry, leading to jamming or repeated failed insertions.
    \item \textbf{Push-Tomato.} Failures are usually caused by small localization or approach-angle errors that compound during pushing. Better contact dynamics or closed-loop state estimation may further improve reliability.
    \item \textbf{Pick-Grocery.} Failures mainly reflect limited spatial generalization in clutter. The policy can overfit to frequent grasp locations and struggle when object layouts differ from the demonstration distribution.
\end{itemize}

Overall, JOPAT is strongest when success depends on temporal consistency, object permanence, and robustness to visual dropout. Its main limitation is fine-grained contact geometry, suggesting that future extensions should combine the proposed pixel-track world-action model with explicit 3D sensing, tactile feedback, or contact-aware objectives.

\clearpage

\end{document}